# Impact of Fake News on Social Media Towards Public Users of Different Age Groups

Kahlil bin Abdul Hakim and Sathishkumar Veerappampalayam Easwaramoorthy
School of Engineering and Technology, Sunway University, No. 5, Jalan Universiti, Bandar Sunway, 47500, Selangor Darul Ehsan, Malaysia
(Email: sathishv@sunway.edu.my)

## Abstract

This study examines how fake news affects social media users across a range of age groups and how machine learning (ML) and artificial intelligence (AI) can help reduce the spread of false information. The paper evaluates various machine learning models for their efficacy in identifying and categorizing fake news and examines current trends in the spread of fake news, including deepfake technology. The study assesses four models using a Kaggle dataset: Random Forest, Support Vector Machine (SVM), Neural Networks, and Logistic Regression. The results show that SVM and neural networks perform better than other models, with accuracies of 93.29% and 93.69%, respectively. The study also emphasises how people in the elder age group diminished capacity for critical analysis of news content makes them more susceptible to disinformation. Natural language processing (NLP) and deep learning approaches have the potential to improve the accuracy of false news detection. Biases in AI and ML models and difficulties in identifying information generated by AI continue to be major problems in spite of the developments. The study recommends that datasets be expanded to encompass a wider range of languages and that detection algorithms be continuously improved to keep up with the latest advancements in disinformation tactics. In order to combat fake news and promote an informed and resilient society, this study emphasizes the value of cooperative efforts between AI researchers, social media platforms, and governments.

# 1  Introduction

The widespread use of social media platforms like X (formerly Twitter), Facebook, TikTok, YouTube, QQ, Instagram and many other by public users has been on the rise since 2018. In a recent survey conducted by Watson (2019) published in Statista, over 77% of respondents in Nigeria had indicated that their source for news is from social media. Although less than 35% of adults in the European region consider news through social media is trustworthy, over 50% of its consumers across Europe are consuming news through social media. In Abbas, Ravi, et al. (2024, pp. 1-5) conference paper, it was evident that references to Facebook had a news outlet traffic market share of over 70%. The impact of fake news was even evident in large scale public and social with the most recent U.S. elections as an example, where there were an estimated 70% of Americans feel that the fake news impacted their level of confidence in the government. According to the research by Lee (2020), users of the age of 65 and over are the most vulnerable to accepting fake news at face value, while a part of the users generally has a distrust in mainstream media built over time. In addition, the exploitation of AI and ML in creating deepfake audio, video and multimedia content is concerning. According to Akhtar et al. (2024), deepfake is an output of content, be it in the form of audio, video or multimedia, that is digitally modified and created with the application of AI and deep learning models. This can include manipulating images and video, such as facial manipulation and swapping the mouth movements to match certain speech alterations.

With the advancement in data sciences and the application of artificial intelligence and machine learning models, fake news detection has become more accurate with the availability of various computational capabilities and models to ascertain a certain news feed is legitimate or can be classified as fake. However, the final judgement also relies on the individual's capability to recognise and ascertain that such news feeds received and read is deemed real or fake. The research objective is to determine the best approach for classification and identification of fake news. There are several objectives that were identified as described below:

- To explore the ability to apply machine learning in fake news classification.
- Identify the features of a fake news classification.
- To analyse the available pre-processing techniques that will help in fake news classification.

Fake news is rampantly spreading in social media platforms like Facebook, TikTok, X (formerly Twitter) and WhatsApp that has a potential impact towards public opinions and social cohesion.

How might the application of Artificial Intelligence (AI) and Machine Learning (ML) identify fake news to potentially reduce the impact of it towards social media users through targeted mitigation strategies?

# 2 Literature Review

In the introduction of this paper, the spread of fake news has an impact on the recent U.S. elections because news and bite-sized articles in online media does not acclimate to the real factors had provoked social cohesion. In the article presented Akhtar et al. (2024), the objective is to look into the current trends in fake news. This include the spreading of articles and the use of deepfake videos, which are considered as techniques in creating and detection of fake news. Advancement in mobile handheld devices providing the computing capacity and capability, coupled with advanced applications with built-in AI capabilities, has enabled users to easily create, re-create and distribute multimedia content. The capabilities in the hands of public users with easy-to-use tools empowers non-expert users to produce deepfakes and digitally modified content poses challenges against current computer forensics experts and tools to identify them. There is a constant need for further investigations towards the potential impact and danger that deepfake and fake news has on the users of different age groups. There are many concerns raised surrounding the evolution and accessibility of tools for deepfake generation. This also poses an opportunity to further advance detection frameworks to ensure that such media generated can be mitigated and stopped from widespread. Deepfake poses a threat through the potential manipulation of speech and facial recognition technologies in AI and deep learning. Without proper adoption and detection framework, misinformation may lead towards malicious impact on socio-economic stability as well as public opinions that results in privacy, security and trustworthiness of social media content.

According to the research (Mushtaq et al., 2024) it was profound that Support Vector Machine (SVM) and Naïve Bayes classifiers have been able to perform better against other Machine Learning models when applied onto a large volume dataset on addressing and classifying fake news. The impact of fake news has on individuals poses a threat on the way every one of us think and perceive information. The affect can be seen as a psychological effect, which was also pertinent in the use cases like the US election as evidenced by Abbas, Ravi, et al. (2024, pp. 1-5). Most articles and research papers focus on the application of Deep Learning and Machine Learning techniques to identify fake news and apprehend such news issues.

The application of data sciences specifically in machine learning towards fake news detection is key to reduce the widespread of such misinformation distribution. However, due to the nature of the misinformation being distributed, it could be in the form of text or through audio and video media – like Deepfakes. According to Abbas, Ravi, et al. (2024, pp. 1-5), we can determine that the use of

Logistic Regression and Multinomial Naïve Bayes (NB) are the most suitable fit for a dataset that is more prone to binary classifications.

In another paper by Mushtaq et al. (2024), we can also summarise that it is important that the approach applied to coming up with an effective solution towards the application of data sciences in fake new predictions. The application of Deep Learning approaches like the use of Convolution Neural Network (CNN) where it was able to identify and classify real against fake news. Additionally, the application of CNN, Gaussian NB, Linear SVM and Gaussian SVM as classifiers have been tested on the accuracy of it.

The use of Naïve Bayes and Support Vector Machine (SVM) in classification of fake news is also tested and found to be most effective when ran within a data acquisition in X (formerly Twitter). This was evidenced in an article by Abdullah, et al. (2024), where a supervised learning model was applied and evidenced that a decision tree model produced a 100% accuracy on text-based data and a metadata consistency of 94.54% which outperforms SVM and Naïve Bayes models. As a result, it was concluded that a decision tree classifier is the most effective in X fake news detection.

Based on the literature review conducted, it was also evident that the thematic measures for the evaluation of machine learning models include accuracy, precision, recall and the f1-score. These key measures are important to ensure that the machine learning algorithms applied are effective.

According to Kaur and Ranjan (2024), we can determine that the use of machine learning in fake news detection is of importance because it allows public users to focus more on important tasks like fact-checking and investigative journalism which fake news detection could be done through machine learning. With the application of Machine Learning, it allows for large-scale dataset handling that will benefit the detection and identification of falsified news.

Another thematic area of applying deep learning in fake news detection is the use of NLP – through Natural Language Processing, this allows for ease of detection of falsified information. The process of NLP provides a baseline to effectively evaluate the credibility of the news sources based on its accuracy and references.

Different age groups interpret news and information differently. In a study by Peng et al. (2024), it was profound that older adults have the tendency to fall into the persuasive misinformation of online health misinformation. The use of persuasive keywords to influence ones' belief, attitude and behaviours is commonly referred to as a persuasive strategy. According to Peng et al. (2024), the three main factors that older adults fall into believing fake information are due to information factors, individual factors, and the interplay between these two factors. Older adults may fall into believing misinformation is due to the persuasive methods applied in the news and linguistics where personal emotions and narratives are used in its texts and information.

The rise of Generative Pre-trained Transformer (GPT) like ChatGPT, Gemini and Copilot, has also been seen to impact an individual's ability to determine real versus misinformation. The effective use of GenAI and GPT is much concerning as studies have shown that excessive use of ChatGPT and such GPT-like tools that is widely available and accessible can potentially develop procrastination, causes memory loss, and dampen the academic performances of young generations.

According to the research by Abbas et al. (2024), its study demonstrates the practical implications of the use of GPT for higher institutions, policy makers, instructors and students. One key finding is that the use of GPT is mainly used due to the factor of time constraints faced by individuals to complete their assignments and work.

The ethical use of GPT amongst individuals particularly in the academic area where it is largely consumed by students is alarming. To a certain degree, the limitations of ChatGPT for example, is that the output is not fact checked by students of the information for credibility and accuracy. According to Kastowo et al. (2024) we can determine that the limitation of ChatGPT of its currency of information is up to 2021, and that mostly the provided output are general statements and is not as contextual as it would be received.

There are current applications of fake news detection and automated corrective strategies being developed and applied. In a research paper by Burel et al. (2024), their approach was built upon four distinct process that applied automated corrective strategies to evaluate its performances. It was profound that the corrective Twitter/X bot in attempting to correct misinformation is greatly accurate for systematic corrections on a large-scale implementation with little to no supervision required. However, the experimentation for introducing a Twitter/X bot in this research produced mixed perceptions from the targeted audience. It was found that an automated misinformation correction in real-world implementations has larger constraints.

In the paper by Indiongo (2024), the need for traditional mainstream media to invest in technological solutions and that collaborative efforts among regional and international cooperations to combat and circumvent or minimise the impact of fake news distribution is essential.

According to the paper by Kaur and Ranjan (2024), the effectiveness of machine learning models need to consider the risks of overfitting and the need for hyperparameter fine tuning. Regularisation is essential to overcome overfitting of data. Different machine learning models need to be tested and applied based on the relative use cases. The use of existing data sciences python libraries like NLTK, Keras and TensorFlow allows for the application of machine learning in the classification tasks.

According to a survey conducted by Mahdi and Shati (2024), there are several types of fake news that could be classified in different categories; namely rumours, misinformation, disinformation, hoaxes and clickbait. It was also found that GNN have made significant impact on the effectiveness of fake news detection. The use of GNN allows for identifying intricate patterns that could possibly have been not considered and captured through conventional techniques like SVM or linear regression.

The rapid adoption of social media platforms like X, Facebook, TikTok, YouTube and many more, has changed the way mainstream media information sharing and consumed. According to the research (T, 2024), these platforms are not only the main source for information but is also the root cause for dissemination of fake news and misinformation. It was found that over 70% of false news stories are more likely to be redistributed against those of true in nature.

According to Khan et al. (2024), the impact of fake news in the Pakistani social assumption particularly through social media information dissemination has an impact through the undermining of its democracy, increase in animosity and through encouraged mistrust. In its research, over 68% of the respondents have a moderate to high level of distrust in mainstream media and conventional news sources. This was due to the implications of a political agenda.

Research on the impact of fake news and misinformation on public trust through traditional media outlets reveals that there are research gaps in the contextual and methodological approaches. According to Indiongo (2024), it was found that fake news disseminated through social media has greatly impacted public trust in traditional media outlets. However, the distrust in mainstream media differs from country to country that could be due to the adoption and impact of social media on its users.

According to Arora et al. (2024), we can conclude that the affect of fake news has on social cohesion is great as news travels through social media in a rapid manner, which has impact on the overall psychological health of an individual. It was profound that there is a direct correlation between the relationship of fake news and the sentiments where it substantially impacts mental health. The psychological impact of fake news distributed over social media platforms will have a psychological impact hence, the development of countermeasures to circumvent fake news distribution is crucial.

In another research conducted and published by Baissa et al. (2024), we can determine that fake news emphasises on elements of negativity and unexpectedness, which has a detrimental effect on news values received by public users. This was also evidenced through an example sighted where fake news undermines the trust in climatologists and creates a doubt on scientific models on the impact of climate changes.

The heavy adaptation of AI and ML increasingly in all aspects in technology and social areas. In a research by Mavrogiorgos et al. (2024), it was evident that biasness in Machine Learning is unavoidable or completely eliminated. Even though there are mitigating controls in ensuring an unbiased ML output, it is essential that the entire lifecycle of Machine Learning and modelling is made through in its entirety and not being conducted in isolation.

The majority of use cases and proof of technology in applying Machine Learning and training the models in fake news detection is mainly focused through English language and in some parts Spanish and Chinese. According to the paper by P et al. (2024), we can determine that the lack of available data for certain geo-political scope may have an impact on the overall performance of fake news detection. It was also evident that the neglect of affect in understanding affective geo-political biasness in fake news detection should have considerations in the development of a more inclusive and applies emotional intelligence in the AI methods to be able to be more inclusive across a multi-cultured global reach.

# 3 Methodology

This section outlines the dataset that has been identified and the proposed methodology for applying different machine learning models to determine how fake news can be identified through proper application of classifications.

## 3.1 Identifying the dataset

In order to proof the methodology and applying this to motion, it is crucial to obtain a set of data that potentially would have its authenticity and accuracy. The dataset is being sourced from Kaggle and obtained. The dataset was sourced from [https://www.kaggle.com/datasets/rajatkumar30/fake-news/data](https://www.kaggle.com/datasets/rajatkumar30/fake-news/data) where this dataset has both fake and real news which has labels against it.

## 3.2 Methodology for fake news prediction

Based on the extensive literature reviews, there are many recommended ML approaches towards identifying and classification of fake news. In order to increase the prediction accuracy and efficiency, the dataset sourced which contains fake news and real news was loaded and we conducted a test based on the top four Machine Learning models to evaluate the overall performance.

### 3.2.2 Approach

Figure 1.1 demonstrates the high-level design for the implementation of a structured approach towards the design and evaluation of the applied Machine Learning models to identify and classify fake news versus real news.

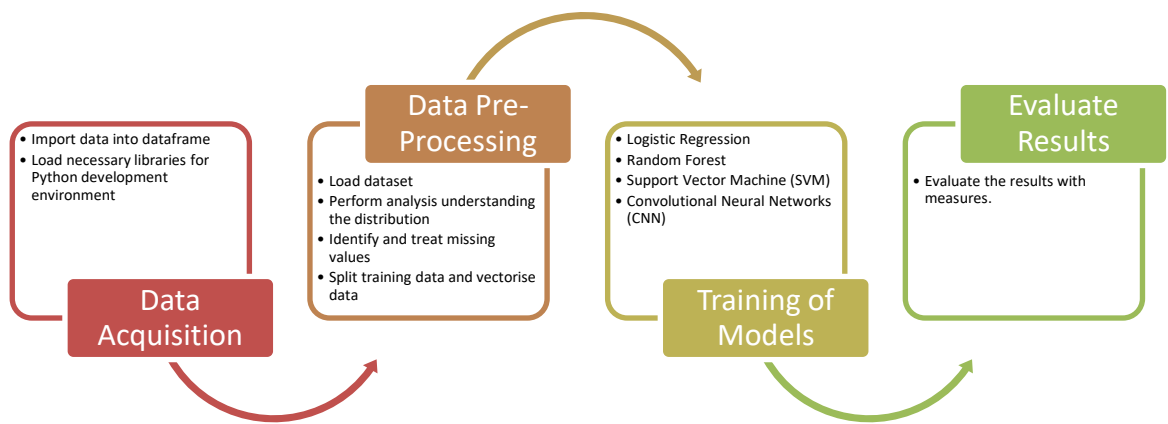

Figure 1.1: An approach to evaluate the most effective machine learning model in identifying and classifying fake news.

### 3.3 Data Acquisition and Data Pre-processing

A development environment is setup with the use of jupyter and the required libraries are being loaded to the development environment. Once the data is loaded to pandas dataframe, basic analysis is performed to identify the distribution of the dataset itself, checking for missing values and to identify the label distributions.

As part of the data pre-processing process, identification of missing values is checked through a df.isnull command and to treat missing values to drop them from the dataset and dataframe. Post handling the data, the data is then encoded to replace the text content where the label column are encoded as binary. In this approach, the binary value is replaced where 1 = real news, and 0 = fake news.

The loaded dataset is split into training and test sets with a test size of 0.2 and a randomiser value of 42. It is also essential and crucial that the text is being vectorised that will allow the machine learning algorithm to handle the text-based content effectively. Vectorising text with TF-IDF provides a more rigid correlation with a human-based assessments. According to Wang (2024), TF-IDF improves the semantic extraction which proves to increase the performance of the models. Term Frequency-Inverse Document Frequency (TF-IDF) is effective as it captures the term importance, reduces the impact of common words in natural language processing (NLP) and creates sparse matrices as it is able to handle large vocabularies and long texts.

An output from the initial data load and pre-processing provides an outcome where there are no missing values identified in the dataset and that the distribution of real against fake news is almost an equal split of 50:50.

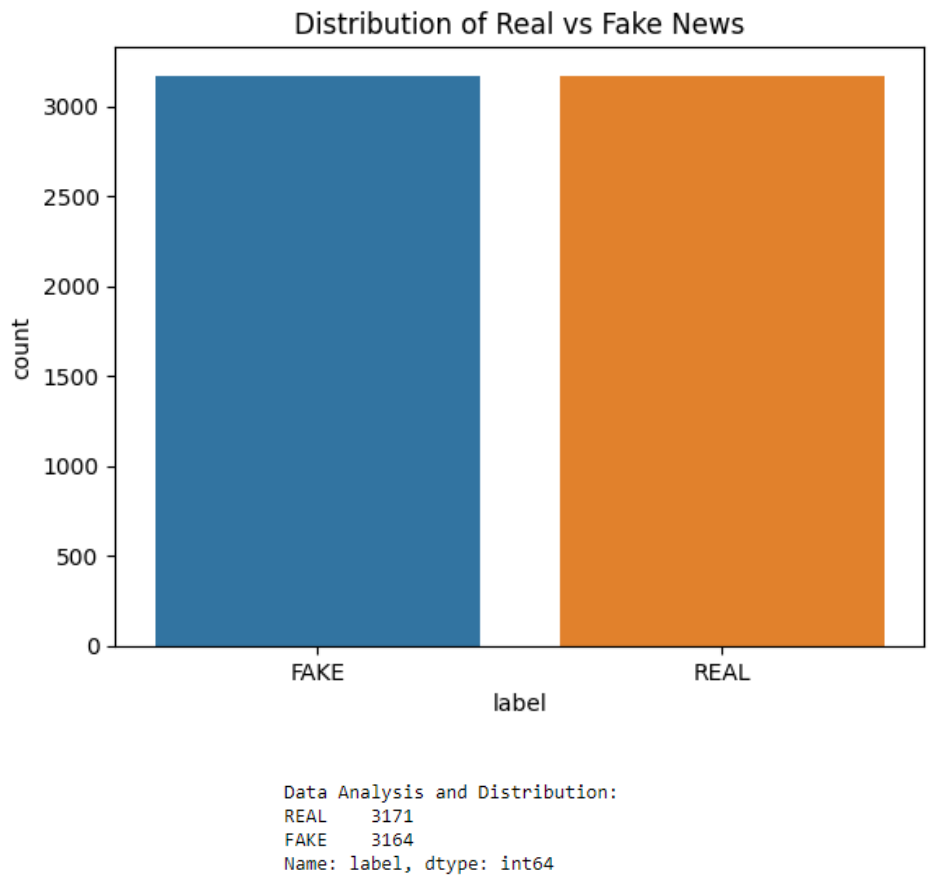

Figure 1.2: Distribution of Real vs Fake news.

## 3.4 Model training

The following four (4) machine learning models were evaluated as part of the evaluation process to determine the most effective and accurate model that will perform best for classifying fake versus real news.

The logistic regression model is chosen as it applies a more linear approach towards text-based classification tasks. Then we move into the application of random forests where the use of a decision tree model tends to perform better in increasing prediction accuracy.

To further evaluate the proposed models as described [2.1], Support Vector Machine (SVM) and Neural Network is also applied in the model training and development. The selection of these four models are based on the available Machine Learning models to further evaluate each and every one of its accuracy and performance.

Initial test in the model training with the use of Neural Network model fell short of its performance through the outputs analysed and evaluated. This could be due to overfitting in the Neural Network. As a result, further fine-tuning was conducted to include regularisation and dropout evaluation to the Neural Networks. Techniques like identifying dropout layers to randomly disable a portion of neurons during training is to prevent the network from depending too much on specific neurons. Regularisation adds weighted penalties to the network to discourage over-complex models.

These machine learning models were chosen as part of the fake news detection is because of their characteristics and ability to handle text classification tasks that are identified in the fake news dataset.

Logistic regression model is a more linear approach that is simple to implement and easier to interpret. In the use case and model training and development, the TF-IDF vectors of the text provide an affect to the probability of the news article being correctly and accurately classified. As a result, the conversion of the binary classification tasks in the dataset suites perfectly well with a logistics regression output and capability to apply a binary classification model. Based on the literature reviews, it was also prodound that the advantages of a logistic regression model is observed in maintaining an efficient computational workload that is faster to train and provides a better performance on a linearly separable dataset.

The random forest on the other hand is a non-linear classification model that leverages on an ensemble of decision trees that can model a non-linear relationship between the features and labels. Based on the dataset that was gathered, this model is useful where patterns of textual data may not necessarily be linear. According to Abdullah, et al. (2024) and based on the output models tested, it was evident that a random forest model can work perfectly well in handling a more complex dataset. The key advantage of a random forest model is that it averages the results through the bagging of many trees, that potentially reduces the risk of overfitting and may provide a more generalised model.

Based on the dataset, the multi-dimensional dataset and pre-processed binary classification tasks in the model building provides a better platform for a support vector machine (SVM) model. A linear SVM provides a solid performance on the TF-IDF vectors that was done in the pre-processing tasks, and can outperform other models known through its robustness and the ability to generalise the dataset. This was also evidenced in a research by Suresh and Kattupalli (2024), where the ability of SVM to find an optimal hyperplane maximises the margin between its classes. This then provides a better generalisation of unseen data, thus making it less susceptible to overfitting compared to other models.

Another machine learning model that was considered in the model development and evaluation of its effectiveness towards fake news detection is the use of neural networks. Similarly to SVM, neural networks is able to handle complex patterns and non-linear relationships like in random forests. These are important tasks and patterns that will increase the dependencies to complete tasks like fake news detection. With the combination of regularisation techniques, neural networks becomes a more flexible and adaptable model compared to other machine learning models. With its capability to handle large-scale datasents, neural networks can learn intricate relationships based on the research demonstrated in Wang (2024) and as an outcome from the literature review according to Abbas, Ravi, et al. (2024, pp. 1-5).

# 4 Results and Discussion

Further analysis was conducted post implementation of the recommended machine learning models to predict and classify fake news. As brought up in Figure 1.2, the distribution of the data is between fake news and real news, with a near equal split of 50:50. There were no missing data from the dataset identified i.e. no null values, which means that the data itself was pre-processed to ensure no missing values are being populated.

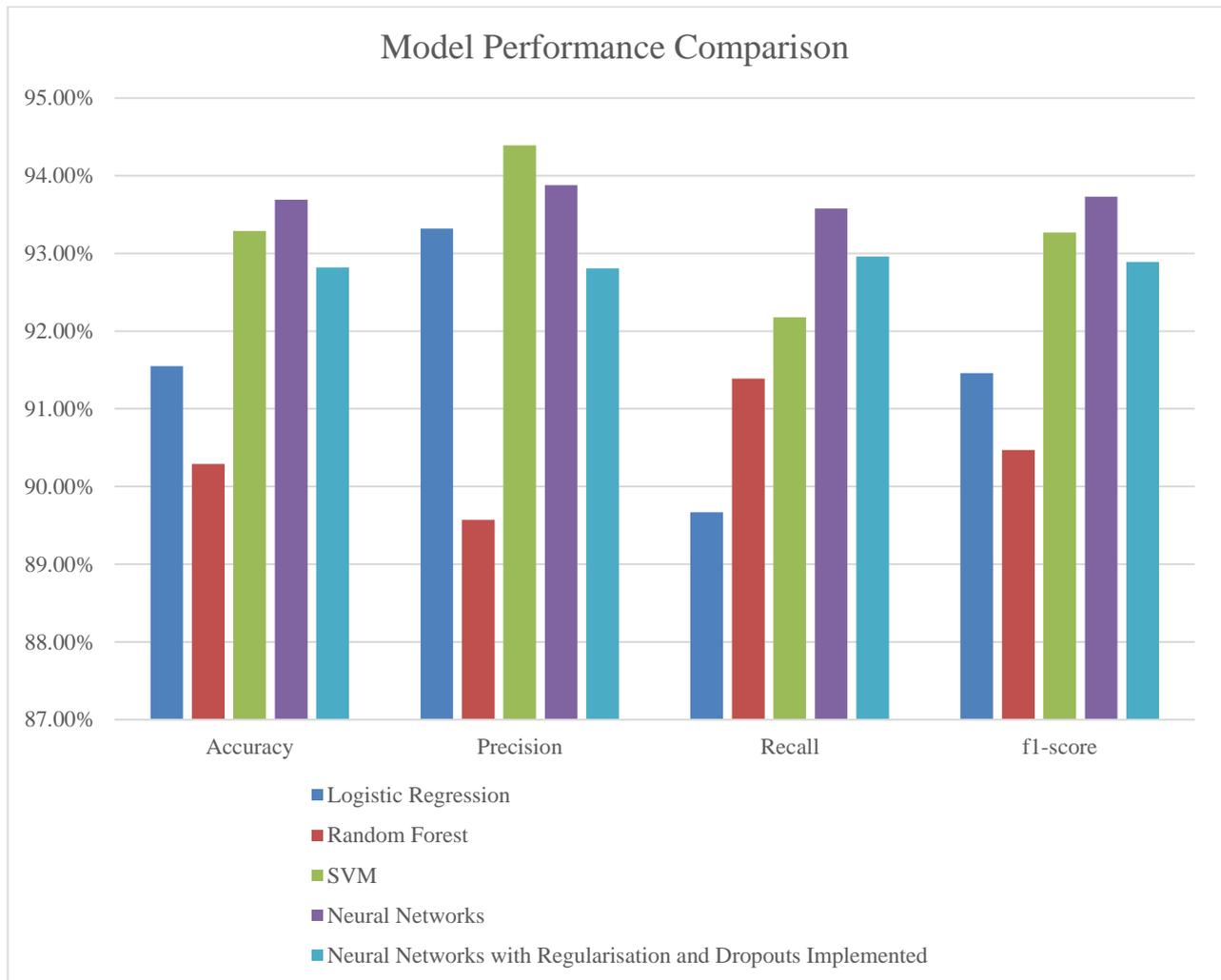

Figure 1.3: Results performance compared between different machine learning models.

In the initial test conducted where neural networks were applied without the use of regularisation and dropouts, neural networks provided the best accuracy of 93.69% with SVM being second highest with 93.29% which makes it best performing in making accurate predictions.

SVM and Neural Networks provide a more consistent result particularly in tasks that requires a good balance between precision and recall. These two models are best suited for high-dimensional texts like fake news detection and is effective in minimising wrong classifications.

| Model | Accuracy | Precision | Recall | f1-score |
|---|---|---|---|---|
| Logistic Regression | 91.55% | 93.32% | 89.67% | 91.46% |
| Random Forest | 90.29% | 89.57% | 91.39% | 90.47% |
| SVM | 93.29% | 94.39% | 92.18% | 93.27% |
| Neural Networks | 93.69% | 93.88% | 93.58% | 93.73% |
| Neural Networks with Regularisation and Dropouts Implemented | 92.82% | 92.81% | 92.96% | 92.89% |

Table 1: Consolidated results from model execution.

As presented in Table 1, logistic regression demonstrates a well performing model in fake news classification based on the dataset available. With a 93.32% precision maintained, it can also be a good choice for fake news classifications. This is essentially valuable when compute resources are limited as logistic regression models are more linear in prospect and a much simpler model approach in machine learning.

Overall performance of the random forest model demonstrates that it is not as effective as the other models due to the low accuracy and low precision. It provides a lesser competitive advantage towards fake news classification which may lead to miss-hits in circumventing fake news spreading widely over social media.

An updated model in the neural networks with the application of data regularisation and dropouts creates a more linear balance between precision and recall. Regularisation prevents overfitting of the data that discourages the model from fitting too closely to the training data. Due to the balanced learning, this allows a better outcome when identifying both fake and real news which leads to a higher f1-score with a more balanced performance across different evaluation metrics.

# 5   Conclusion and Future Work

The research on the impact of fake news widespread through social media towards social cohesion of different age groups provided a baseline on the applicable machine learning and artificial intelligence capabilities that could be leveraged to identifying and classifying fake news. Fake news detection is a demand and should be implemented to circumvent such spreading does not impact a readers' social cohesion and potentially have a negative impact on our psychological state.

It was found and evident that these days, the reliance of social media platforms like Facebook, X, TikTok, WhatsApp and others is increasing. Social public are highly dependent on such news outlets and these news sources has evidently led towards the widespread of uncontrolled and unfiltered news, of such a wide range are considered fake news. The uncontrolled widespread and non-availability of automated tools available a the disposal of users lead towards a potential decrease in public trust, which could potentially impact a nation's economic and political stability, as well as perception of individuals.

Through a thorough identification of machine learning models, with thematic agreements that most effective machine learning models for fake news identification and classification can be drawn to Neural Network and Support Vector Machine (SVM) models. The results show that with a Neural Network model, it demonstrates a marginally higher accuracy compared to SVM whereby the Neural Network model outperforms with an accuracy of 93.69% compared to 93.29% from SVM.

With the application of a more regularised data synthesisation, as well as configuring the dropouts in the neural network proved to balance out the performance results. This is a critical step to prevent overfitting and enhancing the ability of the machine learning model to handle a more complex and diverse set of data.

Although in the literature reviews found that logistics regression can provide an accurate results in classification of fake news versus real news, it was found to fall short in comparison to SVM and neural networks. It is effective in terms of its accuracy, though has the lowest score in recall score and a rather low f1-score metrics.

Older adults are deemed to be susceptible to misinterpreting news, which in most studies, are more prone to being tricked into believing fake news. This is primarily due to their decreased ability to critically evaluate the credibility of news that is widespread through social media. Use of persuasive

keywords in fake news can change the emotional intelligence and charged narrative that are key factors in contributing to the spread of misinformation among older adults.

The application of deep learning and AI technologies in combatting deepfake, a more sophisticated form of fake news is also advanced. With the ability to integrate the models with natural language processing (NLP), deep learning techniques may be a promising architecture for enhancing the accuracy and scalability of fake news detections.

One key area for potential future work is the ability to process and expand the datasets to include non-English or prominent languages. This will help in covering a wider and diverse geopolitical context that will significantly cover fake news detection globally. With the advancement in generative AI creating fake news and deepfake videos, it is crucial that continuous work and refinement in the detection algorithms are being kept up to date to ensure such techniques can cover a more vast coverage of fake news detection.

In conclusion, the impact of fake news through social media platforms is particularly vulnerable to different and targeted age groups. This poses a serious threat towards the overall social cohesion and the integrity of information. There are AI and ML models present that provides tools towards reducing the widespread of fake news. However, the ever-evolving landscape of technological advancements, fake news detection may become more harder to trace without continuous improvements in the machine learning and application of NLP in the models training. The growth of AI-generated content, like GPT4 in ChatGPT, poses a threat as the generation of fake news becomes more accessible and can be made more targeted towards susceptible target audience in the older generation. The effectiveness of combatting fake news widespread over social media can be reduced and mitigated through continuous development and crowd-sourcing of integrating machine learning models as well as through continuous education.